\documentclass[letterpaper, 10 pt, conference]{ieeeconf}  

\IEEEoverridecommandlockouts                              

\usepackage{graphicx}
\usepackage{amsmath} 
\usepackage{amssymb}  
\usepackage{hyperref}
\usepackage{wasysym} 
\usepackage{ulem} 
\usepackage{arydshln} 
\usepackage{blindtext}

\usepackage{xcolor}

\usepackage{multirow}

\definecolor{supergood}{rgb}{.2,.6,.2} 
\definecolor{good}{rgb}{.5,.99,.5} 
\definecolor{equal}{rgb}{0.1,0.1,0.1}  
\definecolor{bad}{rgb}{.9,.7,.1} 
\definecolor{superbad}{rgb}{.9,.1,.1}   
\definecolor{notworking}{rgb}{.5,.5,.5}

\title{\LARGE \bf
\Large What makes visual place recognition easy or hard?
}

\author{
Stefan Schubert and Peer Neubert \\ 
TU Chemnitz, Germany\\
\texttt{ \{stefan.schubert,peer.neubert\}@etit.tu-chemnitz.de} \\
}

\begin{document}

\maketitle
\thispagestyle{empty}
\pagestyle{empty}

\begin{abstract}

Visual place recognition is a fundamental capability for the localization of mobile robots. 
It places image retrieval in the practical context of physical agents operating in a physical world.
It is an active field of research and many different approaches have been proposed and evaluated in many different experiments.
In the following, we argue that due to variations of this practical context and individual design decisions, place recognition experiments are barely comparable across different papers and that there is a variety of properties that can change from one experiment to another. 
We provide an extensive list of such properties and give examples how they can be used to setup a place recognition experiment easier or harder.
This might be interesting for different involved parties: 
(1) people who just want to select a place recognition approach that is suitable for the properties of their particular task at hand,
(2) researchers that look for open research questions and are interested in particularly difficult instances, 
(3) authors that want to create reproducible papers on this topic, and
(4) also reviewers that have the task to identify potential problems in papers under review.

\end{abstract}

\section{Introduction}

The task of visual place recognition (VPR) is to find associations between one or multiple query images with a database of images of known places.
It is an important means for mobile robot localization, in particular for loop-closure detection during SLAM or for candidate selection for pose estimation.
Due to its importance for long-term operation, a particularly active research field is recognizing places despite severe changes of the environmental conditions induced by day-night cycles, changing weather, or seasons. 
At its core, VPR is an image retrieval problem and many developments and results from the computer vision community (e.g. image descriptors) significantly contributed to the state of the art.
However, the application in the context of mobile robotics frames the image retrieval task and typically provides additional structure and information (e.g. spatio-temporal consistency).
This framing had two important effects for the research in this area:
First, adapting to the different framing led to the emergence of own datasets and evaluation best-practices in the robotics community which are different to the computer vision community. In particular, the standard datasets and benchmarks from computer vision are typically not applied. 
Second, dependent on the additional structure and information that is available and exploited, significantly different types of place recognition problem setups emerged.
This leads to a huge variety and volatility of experimental setups and evaluation procedures among different papers.
Which then leads to barely comparable results between these papers and a lack of common understanding of the state of the art. 
The negative consequences include a fragmentation of the community and severe impediments for monitoring the advances in the field and for identification and structured addressing of open research questions.

In this article, we provide an overview of the different existing place recognition problem setups. Starting point is a presentation of a basic place recognition pipeline in Sec.~\ref{sec:pr_basics}. This is used to demonstrate the influence of design decisions on the comparability of results across groups and papers in Sec.~\ref{sec:barely_comparable}.
We will then provide an extensive list of properties of place recognition problems that can have large influence on the difficulty of the task in Sec.~\ref{sec:pr_props}.
Finally, we will identify particularly challenging problem setups and property combinations that are interesting for future research in Sec.~\ref{sec:discussion}.

\section{Preliminaries: A basic visual place recognition (evaluation) pipeline}
\label{sec:pr_basics}

\begin{figure}[b]
    \centering
    \includegraphics[width=1\linewidth]{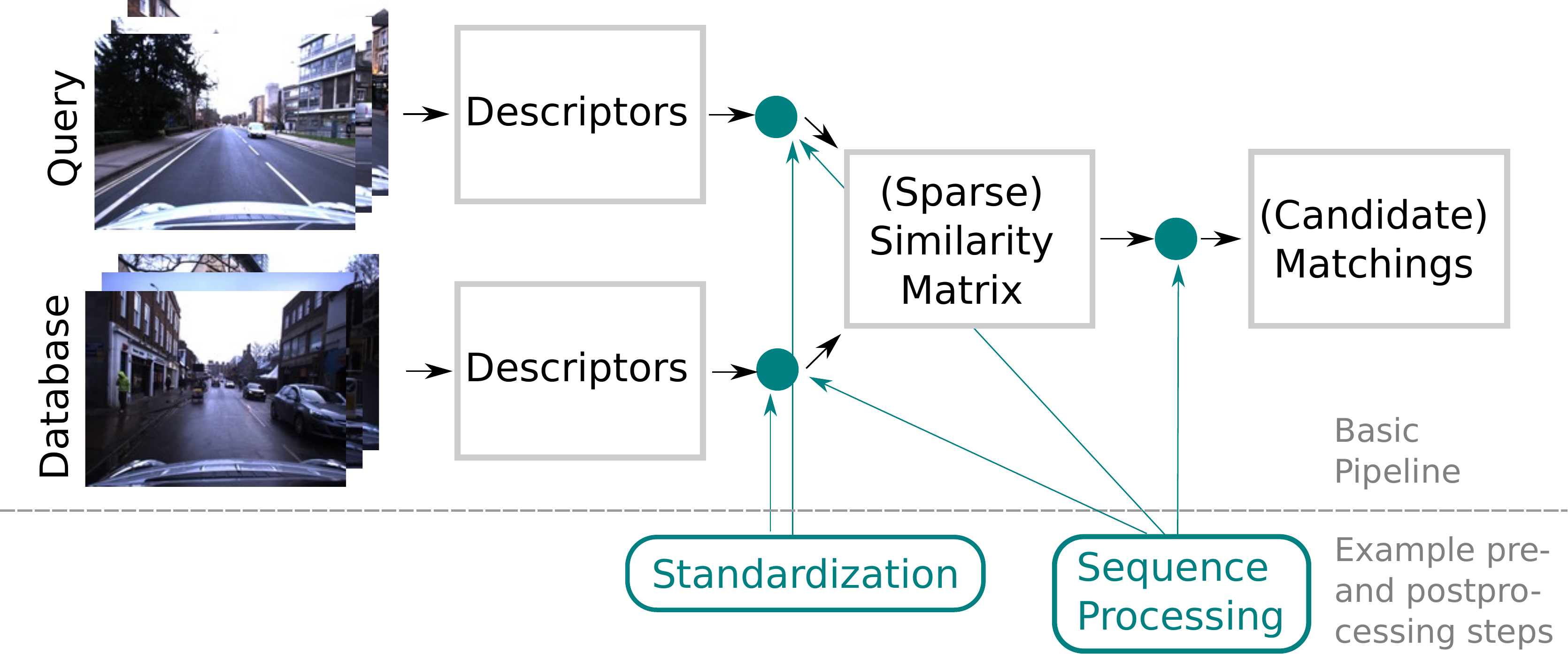}
    \caption{Even this very basic place recognition pipeline with descriptor and similarity matrix computation has many degrees of freedom that need to be specified. This becomes even more complex if it is extended with pre- and postprocessing steps.}
    \label{fig:infl_prepost}
\end{figure}

Visual place recognition is an important subproblem of mobile robot localization and subject of intense research, please refer to \cite{Lowry2016a,Masone21} for surveys.
It is required for tasks like loop closure detection in SLAM or candidate selection for global localization.
Fig.~\ref{fig:infl_prepost} shows a basic place recognition pipeline.
Input are database and query sets, the output is a set of (candidate) matchings between the two sets.
The basic computing steps include (1) computation of descriptors, (2) pairwise comparison of image descriptors to generate a (potentially sparse) similarity matrix and (3) potentially the deduction of a set of discrete matchings (or matching candidates) between the two image sets.
The basic pipeline can be extended with additional pre- and postprocessing steps, e.g. for descriptor standardization \cite{Schubert2020} or sequence processing \cite{Neubert2019}.

However, not all place recognition approaches follow this structure. For example, end-to-end deep-learned methods (e.g. PoseNet-like approaches~\cite{posenet}) encode information about the database in their neural network weights and infer the image pose from a single query image without any explicit descriptor or image comparison. 

Nevertheless, we want to use this basic pipeline to discuss the comparability of place recognition experiments and to identify their distinguishing properties. More details on the individual parts of this basic pipeline can be found in Appendix A. If Fig.~2 makes sense to you, there is probably no need to read this appendix, otherwise, it might be helpful to have a look.

\begin{table*}[tb]
\centering
\caption{A selection of datasets that have been used for place recognition in mobile robotics.}
\resizebox{\textwidth}{!}{%
\begin{tabular}{lc p{6cm} ccc}
	\textbf{Dataset}         &           \textbf{Ref}            & \textbf{Appearance Change}                     & \textbf{Setup} & \textbf{\#Seq} &      \textbf{GT}     \\ 
	\hline
	Alderley Day/Night       &        \cite{seqSLAM}         & day, night/rain                                & car            &       2        &           ID           \\
	CMU                      &           \cite{ds_cmu}           & time, 4-seasons                                & car            &       16       &          raw GPS           \\
	Freiburg Across Seasons  &    \cite{ds_freiburg_seasons}     & summer/sunny, winter/low-sun, long-term change & car            &       3        &           ID           \\
	Freiburg-Bonn            &      \cite{ds_freiburg_bonn}      & summer-winter, morning-evening                 & car            &      2x2       &           ID           \\
	Gardenspoint Walking     & \cite{ds_gardenspoint_riverside}  & day-night, viewpoint                           & hand-held      &       3        &           ID           \\
	NCLT                     &          \cite{ds_nclt}           & sunny/cloudy, time, seasons                    & robot          &       27       &         poses          \\
	Nordland                 &        \cite{ds_nordland}         & seasons, no viewpoint                          & train          &       4        &           ID           \\
	Oxford RobotCar                  &        \cite{ds_robotcar}         & time, weather, seasons                         & car            &      $>$100    &          raw GPS         \\
	SFU Mountain Dataset     &           \cite{ds_sfu}           & time, day-night, weather, seasons              & robot          &       7        &           ID           \\
	South Bank Bicycle       & \cite{ds_gardenspoint_riverside}  & day-night, viewpoint                           & bicycle        &       2        &           ID           \\
	StLucia                  &         \cite{ds_stlucia}         & time of day                                    & car            &       10       &          raw GPS           \\
	Symphony Lake            &      \cite{ds_symphony_lake}      & illumination, seasons                          & boat           &      121       &         poses          \\
	V4RL Urban Place         &          \cite{ds_v4rl}           & viewpoint                                      & hand-held      &       3        &           ID           \\ 

	All day                  &         \cite{ds_all_day}         & day-night, time, dusk, dawn                    & car            &       6        &          raw GPS           \\
	City Center              & \cite{ds_city_center_new_college} & dynamics, shadows                              & robot          &       1        &           ID           \\
	DIRD                     &          \cite{ds_dird}           & time                                           & car            &       2        &     not available \\
	Ford Campus              &          \cite{ds_ford}           & sunny, overcast                                & car            &      2x1       &          raw GPS           \\
	Kelvin Grove Footpath    &         \cite{ds_kelvin}          & day-night                                      & hand-held      &       2        &           ID           \\
	Kitti Odometry           &          \cite{ds_kitti}          & dynamics                                       & car            &       11       &         poses          \\
	Malaga Urban Dataset     &         \cite{ds_malaga}          & dynamics                                       & car            &        1       &          raw GPS           \\
	Mapilary Berlin          &        \cite{ds_mapilary}         & dynamics, viewpoint                            & bicycle, bus   &       2        &          raw GPS           \\
	New College              & \cite{ds_city_center_new_college} & dynamics                                       & robot          &       1        &           ID           \\
        \"Orebro Seasons         &         \cite{ds_seasons}         & seasons, weather                               & robot          &       7        &         poses            \\
	PitOrlManh               &       \cite{ds_pitorlmanh}        & dynamics, 3 cities                             & street view    &      3x1       &         poses          \\
	RAWSEEDS                 &        \cite{ds_rawseeds}         & time                                           & robot          &      12+5      &       poses       \\
	StLucia Vision Dataset   &     \cite{ds_stlucia_vision}      & none (no loop)                                 & car            &       1        &          raw GPS           \\
	Surfers Paradise         &         \cite{ds_surfer}          & day/rain-night                                 & car            &       2        &           ID           \\ 

\end{tabular}%
}

\label{tab:datasets}
\end{table*}

\begin{figure}[t]
    \centering
    \includegraphics[width=1\linewidth]{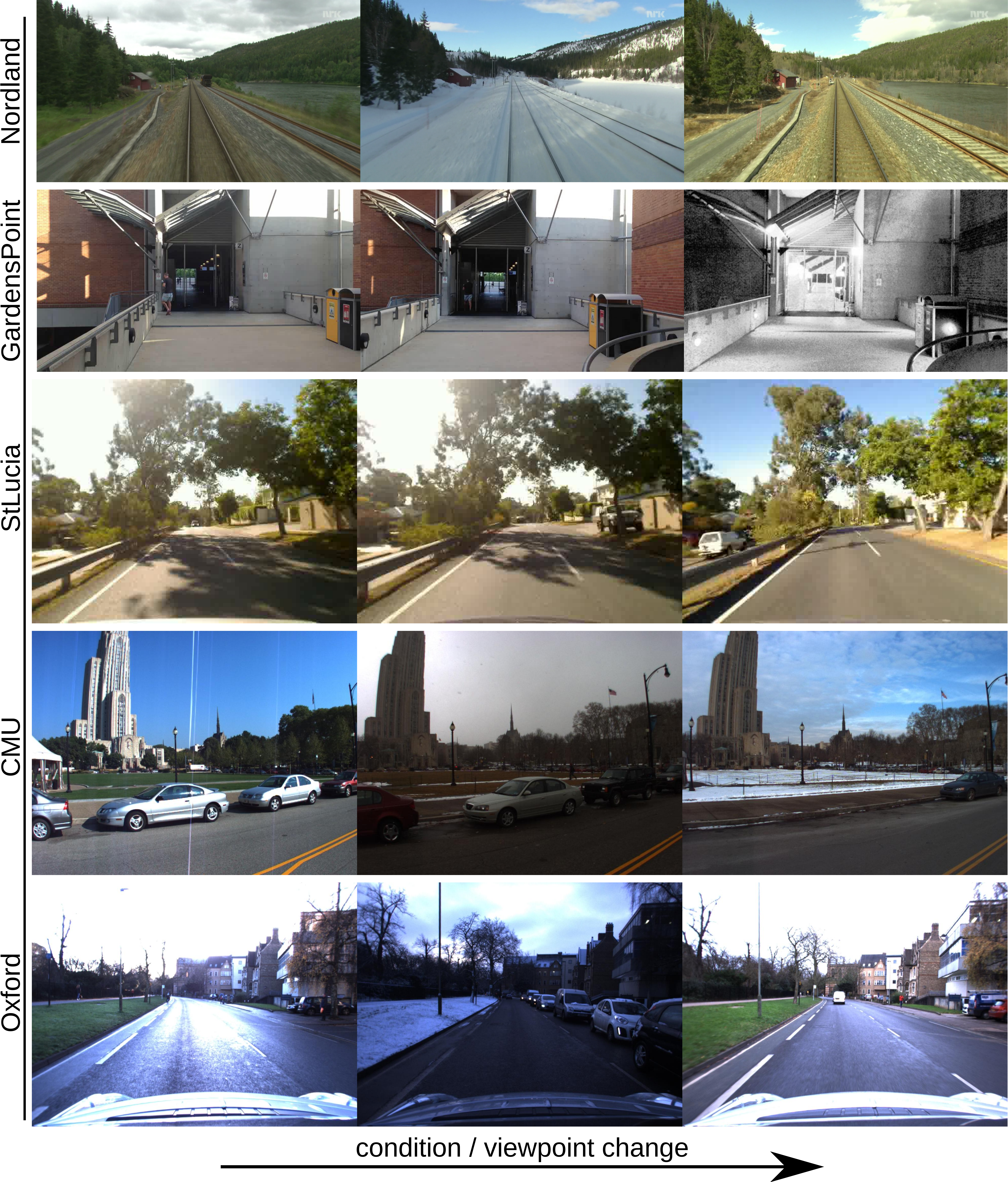}
    \caption{Example images of same places under different conditions for five standard datasets.}
    \label{fig:datasets}
\end{figure}

Foundation for performance evaluation and comparison of place recognition approaches is ground-truth knowledge about matchings between database and query images.
Similar to other robotics applications, the generation of datasets with ground-truth information is a laborious task.
However, in case of place recognition, we face an additional (and potentially more severe) problem:
Place recognition is an ill-defined problem and there is no inherently correct or wrong answer to the question whether two images show the same place or not.
For example, an image of the Eiffel tower and another image of Notre-Dame cathedral can be considered to show \textit{different} places within the city of Paris, or to show the \textit{same} place (``Paris'') within France.
To account for this problem, the creation of ground-truth information for place recognition (or more precisely, creation of the target output of the place recognition algorithm), typically adds additional constraints to the problem, e.g. on the maximum allowed spatial distance of camera poses or difference between image indexes.

Table.~\ref{tab:datasets} shows a collection of datasets that have been used for the evaluation of place recognition algorithms, Fig.~\ref{fig:datasets} shows some example images.
They provide a large variety of properties of database and query sets including the type of environment (e.g., indoor/outdoor, urban/rural), type of condition change between database and query (e.g., dynamics, weather, seasonal), constant conditions \textit{within} database and query or not, viewpoint changes, minimum/maximum spatial distance between consecutive images (e.g., with/without stops), loops within database or query, and so on.
Many datasets are publicly available and were repeatedly used in different works. However, due to the complexity of some of the datasets and potential ambiguities in the ground-truth place matchings, the comparison of results on a (at first glance) same dataset from different papers is unreliable. This will be discussed in more detail in the following sections.

Due to the large variety of potential environments and environmental conditions, it is good practice to evaluate algorithms on a combination of available datasets. 
However, this choice opens opportunities to cherry-pick datasets where the preferred algorithm performs best. There have been a few attempts to provide a systematic collection of datasets and to establish a standard benchmark. They are shortly reviewed in the Appendix B.
However, so far there is no established standard for the evaluation of place recognition algorithms. This leads to the fundamental problem which is discussed in the next section.

\section{Place recognition results are barely comparable across papers}
\label{sec:barely_comparable}

How can I select the best place recognition approach for my task at hand? Can I just look at the most recent papers and choose the algorithm with the best performance measure? Unfortunately, it's not that simple. In this section, we show some examples, where differences in the experimental setups between different papers can have a huge influence on the results and make comparisons across papers largely impossible. The most obvious is the previously mentioned choice of datasets, but there are others, more subtle influences.
The following Sec.~\ref{sec:pr_props} will then provide a more extensive list of properties that allow to distinguish different place recognition problems and that have to be taken care of to avoid such differences.

\subsection{Choice of the dataset}\label{sec:problem_dataset}
Table~\ref{tab:datasets} shows a selection of datasets that have been used for mobile robot place recognition experiments.
Of course, the performance of an algorithm can significantly vary between datasets. 
However, and very importantly, also the relative performance of two algorithms can switch between different datasets. Often, different algorithms will perform best on different datasets.

For example, table~\ref{tab:influence_dataset} shows results presented in \cite{Schubert2020} of the AlexNet-based image descriptor from \cite{Suenderhauf2015} and NetVLAD \cite{netvlad} descriptors  using a simple pairwise image comparison on the Oxford \cite{ds_robotcar} and Nordland \cite{ds_nordland} datasets. The performance of the raw and the standardized NetVLAD descriptors is significantly better for each comparison from Oxford. In contrast, for each Nordland comparison the AlexNet-based descriptor is better.
With knowledge about the details of the algorithms and datasets, we can speculate about the reasons: The constant viewpoint in Nordland presumably helps the flattened-feature-map-style AlexNet descriptor that includes a lot of spatial information; and the domain of the training data for NetVLAD (Pittsburgh 250k) is significantly more similar to Oxford than to Nordland. 
This example demonstrates that the choice of evaluation datasets can considerably influence the result. This allows to (unintentionally) cherry-pick datasets in favour of a particular algorithm.

\boldmath
\begin{table}[tb]
\centering
\caption{Evaluation of the average precision of AlexNet and NetVLAD descriptors on Nordland and Oxford datasets, either using raw descriptors or with standardization \cite{Schubert2020}. The colored arrows indicate large ($\ge$25\%) or medium ($\ge$10\%) deviation from ``Raw''. See \cite{Schubert2020} for details.}
\resizebox{0.45\textwidth}{!}{%
  \begin{tabular}{lll|ll|ll}
  \hline
                   &             &                 & \multicolumn{2}{c|}{\textbf{NetVLAD}}             & \multicolumn{2}{c}{\textbf{AlexNet}} \\
\textbf{Dataset} & \textbf{Reference} & \textbf{Query}  & \textbf{Raw}  & \textbf{STD}  & \textbf{Raw}  & \textbf{STD} \\
\hline       
Nordland & fall & spring & 0.39 \color{equal}{} &  0.61 {\color{supergood}{$\uparrow$}} &  0.82 \color{equal}{} &  \textbf{0.92} {\color{good}{$\nearrow$}}  \\
 & fall & winter & 0.06 \color{equal}{} &  0.26 {\color{supergood}{$\uparrow$}} &  0.63 \color{equal}{} &  \textbf{0.83} {\color{supergood}{$\uparrow$}}  \\
 & spring & winter & 0.11 \color{equal}{} &  0.37 {\color{supergood}{$\uparrow$}} &  0.59 \color{equal}{} &  \textbf{0.86} {\color{supergood}{$\uparrow$}}  \\
 & summer & spring & 0.32 \color{equal}{} &  0.58 {\color{supergood}{$\uparrow$}} &  0.77 \color{equal}{} &  \textbf{0.90} {\color{good}{$\nearrow$}}  \\
 & summer & fall & 0.63 \color{equal}{} &  0.84 {\color{supergood}{$\uparrow$}} &  0.94 \color{equal}{} &  \textbf{0.97} \color{equal}{$\rightarrow$}  \\
Oxford & 141209 & 141216 & 0.87 \color{equal}{} &  \textbf{0.92} \color{equal}{$\rightarrow$} &  0.49 \color{equal}{} &  0.65 {\color{supergood}{$\uparrow$}}  \\
 & 141209 & 150203 & 0.93 \color{equal}{} &  \textbf{0.96} \color{equal}{$\rightarrow$} &  0.63 \color{equal}{} &  0.85 {\color{supergood}{$\uparrow$}}  \\
 & 141209 & 150519 & 0.83 \color{equal}{} &  \textbf{0.91} \color{equal}{$\rightarrow$} &  0.25 \color{equal}{} &  0.80 {\color{supergood}{$\uparrow$}}  \\
 & 150519 & 150203 & 0.85 \color{equal}{} &  \textbf{0.94} {\color{good}{$\nearrow$}} &  0.30 \color{equal}{} &  0.89 {\color{supergood}{$\uparrow$}}  \\

  \hline
  \end{tabular}
}
\label{tab:influence_dataset}
\end{table}
\unboldmath

\subsection{Details of the usage of the particular dataset}

\begin{figure}[t]
 \centering
 \includegraphics[width=0.6\linewidth]{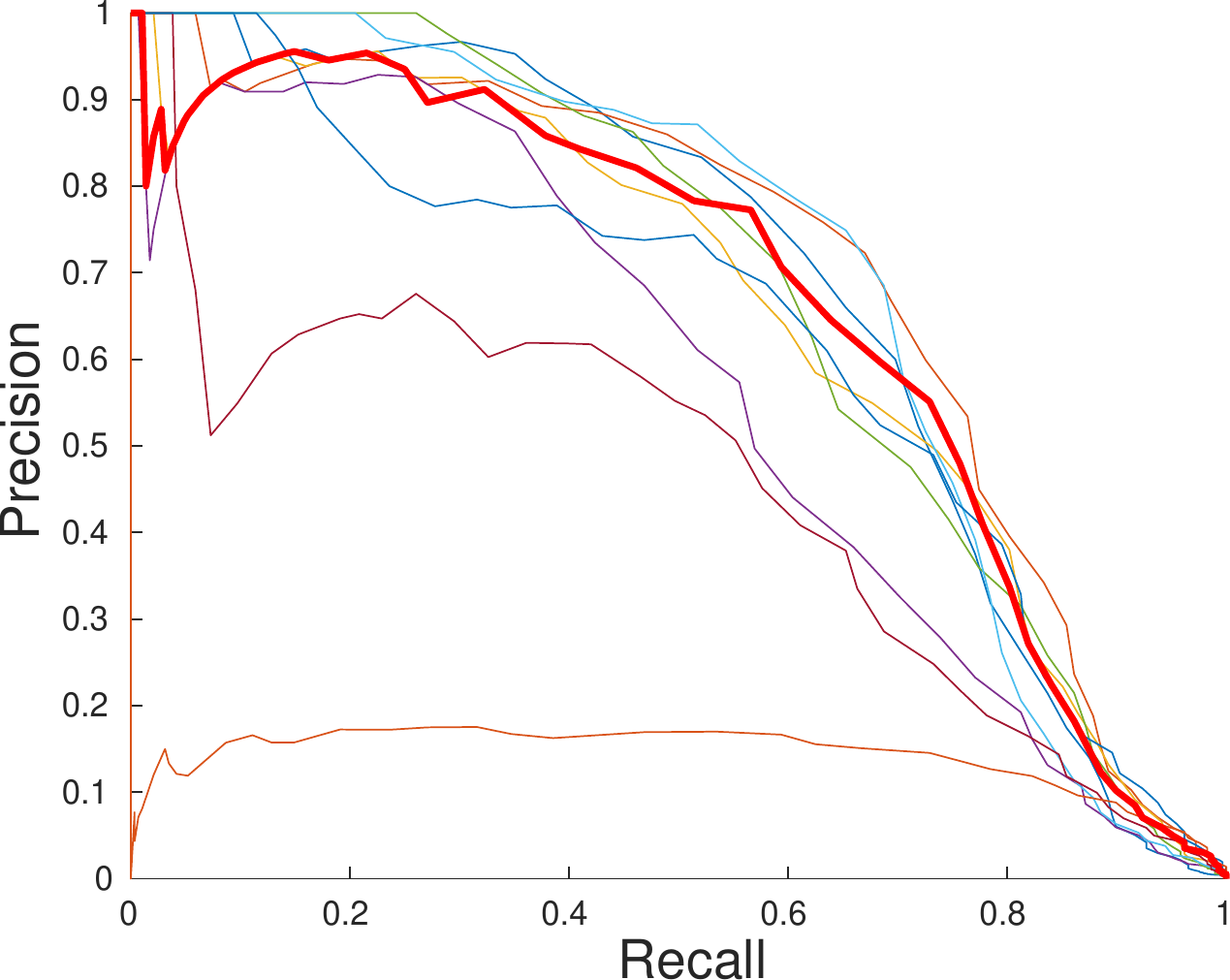}
 \caption{The performance of an algorithm can vary considerably over different fractions of the same dataset. Each curve shows results of the AlexNet on a different fraction of Nordland spring-winter dataset.} 
 \label{fig:infl_data_part}
\end{figure}

Even if the same image source is used, the exact way how a dataset is used can have a huge influence. 
A practically relevant example is to use only a part of a specific dataset.
For instance, Fig.~\ref{fig:infl_data_part} shows precision-recall curves of the \textit{same} algorithm (AlexNet \cite{Suenderhauf2015} pairwise as before) using the \textit{same} basic dataset (Nordland spring-winter \cite{ds_nordland}) but \textit{different fractions} of the trajectory. The thick red curve is obtained when equally sampling images over the whole trajectory. For the others the same number of places is used, but each time sampled only from a smaller part of the trajectory - resulting in a wide variety of curves.
Other examples are differences in the sampling rate of images or whether additional information (e.g., odometry or information about the current condition) is used or not.

\subsection{Single matching constraints}
\begin{figure}[t]
 \centering
 \includegraphics[width=1\linewidth]{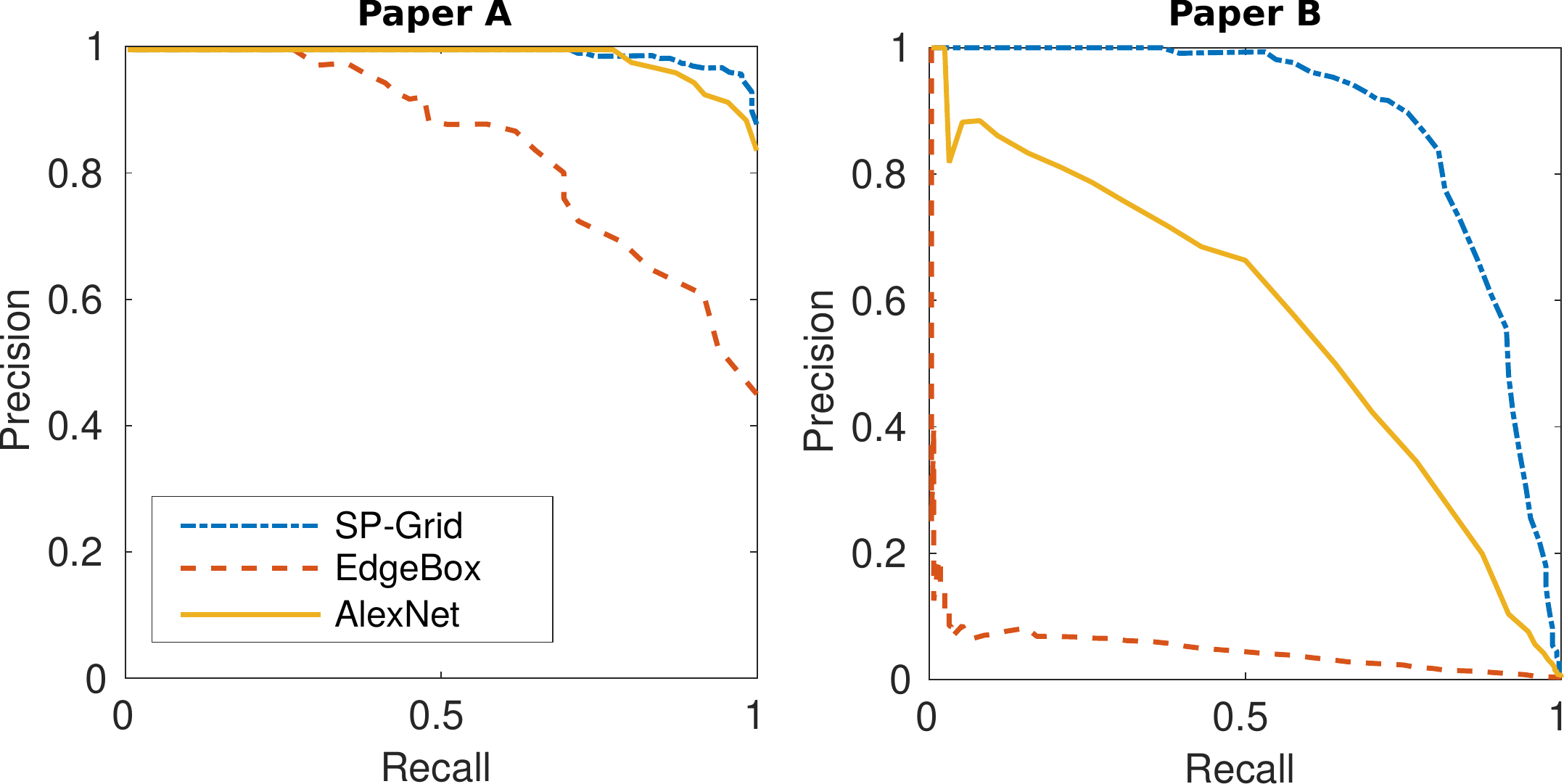}
 \caption{Two sets of curves on the exact same imagery (Nordland spring-winter) created using evaluation procedures from two different papers (A~\cite{ds_mapilary} and B~\cite{Neubert2015}). The same algorithm can show significantly different absolute performance if the evaluation mechanism is changed between papers. Paper A evaluates only the single best matching, paper B evaluates all matchings.} 
 \label{fig:infl_single_match}
\end{figure}

Even if the algorithms are evaluated on the very same image sets, there are other factors that can have huge influence on the performance measures of an algorithm. Most importantly, this can be the measure itself.
For instance, the precision-recall curves in Fig.~\ref{fig:infl_single_match} are recreated according to two research papers (\cite{Neubert2015} and \cite{ds_mapilary}) of two different groups. Both plots show results of several place recognition algorithms on the exact same Nordland spring-winter dataset. Although some of the algorithms (EdgeBox, Holistic CNN, SP-Grid) are evaluated using the procedure from \textit{both} groups their absolute performance is completely \textit{different}. For example, EdgeBox seems to be a reasonable algorithm for place recognition on this dataset in the right plot, but seems to be unsuited according to the left plot. 
Although both plots are precision-recall curves of place recognition on the same dataset, the underlying tasks were significantly different.
The task evaluated in the left plot was to retrieve \textit{all} relevant database images for a query image, while the task for the right plot was to retrieve a \textit{single} relevant database image (this distinction will also be topic of Sec.~\ref{sec:pr_props_output}). Obviously, the two tasks are differently difficult and the curves can not be compared \textit{across} tasks (however, at least in this case, the relative ordering of the algorithms within each task is consistent).

\subsection{Details of the usage of the ground truth}
\label{sec:barely_comparable_gt}
Although the usage of the ground-truth matching information is related to the previous more general discussion on dataset details, we want to emphasize the particular importance and problematic nature of ground-truth information for place recognition.
Place recognition is an ill-defined problem: In general, the information provided by a query and a database image is not sufficient to definitively answer the question whether they show the same place or not - this answer always depends on additional information, e.g., a maximally allowed spatial and angular distance between two camera poses. 

\begin{figure}[t]
 \centering
 \includegraphics[width=0.5\linewidth]{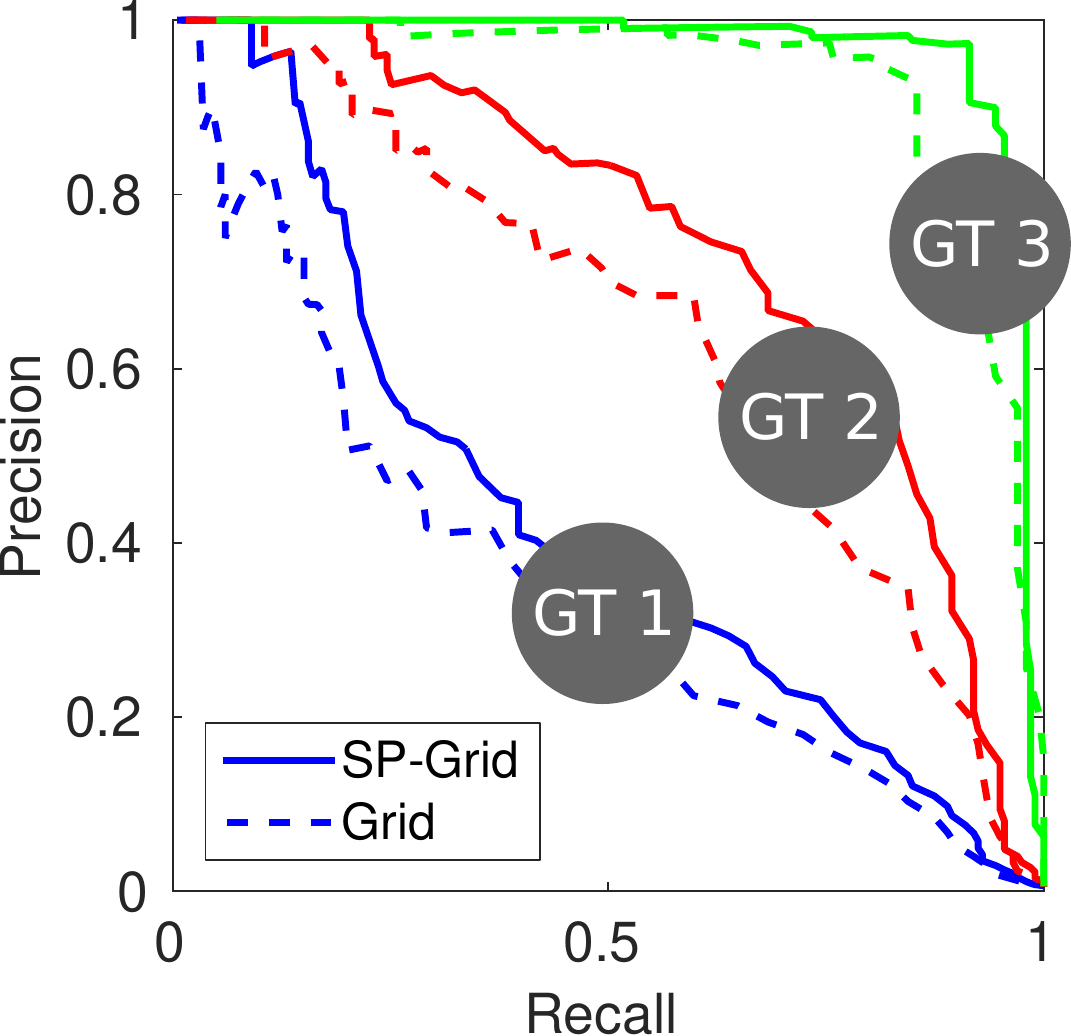}
 \caption{Influence of a varying maximally allowed distance of images to be considered as ground-truth matchings on the Gardenspoint Walking DayLeft-NightRight sequence. The different ground-truths are indicated by GT1/GT2/GT3. Of course, it makes no sense to compare a solid curve (Grid approach) to a dashed curve (SP-Grid approach) if they use different ground truth, which might be the case if they were from different papers.} 
 \label{fig:infl_gt_distance}
\end{figure}

The here discussed visual place recognition problems are typically in the context of mobile robotics, hand-held cameras, or automotive applications. 
For images of the same place, we can expect a considerable overlap in their fields of view, but the particular maximum amount of allowed camera pose displacement for images of the same place can be different.
Many available datasets come either with associated GNSS (e.g., GPS) information or synchronized image sequences.
However, ground truth based on GNSS signals is error prone and requires (manual) correction.
Even without such errors, the choice of the maximum spatial and angular distance in a GNSS pose or the maximum distance of frame indexes of ground-truth matchings significantly influences the overall results, an example is illustrated in Fig.~\ref{fig:infl_gt_distance}.
Most datasets don't come with an obligatory choice of these thresholds, this allows to tailor them to the level of viewpoint invariance that is suitable for a particular task or algorithm. However, variations also prevent a comparison of results between different papers.

\subsection{Pre- and postprocessing steps}
\label{sec:barely_comparable_prepost}
Fig.~\ref{fig:infl_prepost} shows a basic place recognition pipeline with placeholders for descriptor and similarity computation (e.g. simple exhaustive pairwise comparison). This basic pipeline can be extended with pre- and postprocessing steps that can have a large influence on the overall performance. 
Examples for additional steps are the combined evaluation of multiple sequential images (e.g. \cite{Neubert2019,seqSLAM}) or the standardization of descriptors \cite{Schubert2020}. 
The comparison of results for raw and standardized descriptors in table~\ref{tab:influence_dataset} shows examples of the huge possible improvements due to such an  additional simple standardization step; similar results for sequence processing can, e.g., be found in \cite{Neubert2019}. 

When comparing the performance of different place recognition approaches, we have to keep in mind that these pre- and postprocessing steps are largely independent of the particular descriptor or comparison algorithm. 
For instance, a novel descriptor that already includes this descriptor standardization step has an inherent benefit over other descriptors that do not explicitly integrate standardization. Since descriptor standardization can be seen as a default preprocessing step, the novel descriptor should only be compared against existing descriptors \textit{with} the additional standardization step - otherwise, we would not be able to distinguish whether some improvement is due to the novel descriptor computation or merely due the (already known) standardization.

\subsection{Limited transitivity of comparison to baseline algorithms}
A best practice to compare new algorithms to the state of the art is to compare to (a small set of) repeatedly used baseline algorithms (e.g., FAB-MAP~\cite{ds_city_center_new_college} or \mbox{SeqSLAM}~\cite{seqSLAM}).
While this provides some information in comparison to the baseline algorithm, we have to be careful to use the baseline algorithms as a link between different papers and experiments.
A naive approach could be: If all new algorithms are compared against the same baseline algorithm, we can indirectly compare all new algorithms, e.g., `If the existing algorithm X is 10\% better than FAB-MAP and my new algorithm Y is 30\% better than FAB-MAP, then my new algorithm Y is $\sim$18\% better than algorithm X.''. 

Of course, in practice, it is not that simple. Although comparison against a baseline provides some information, differences in the evaluation (i.e., those mentioned in this section) prevent this type of math.
Moreover, if non-optimized parameter settings are used for the baseline algorithms (e.g., vocabulary in FAB-MAP or the sequence length in \mbox{SeqSLAM}) or datasets are used that are particularly challenging for the baseline algorithms, then it might become very easy to beat the baseline by a large margin but with only little information about the relation to the state of the art.
For example, algorithm X could be compared with FAB-MAP on a dataset where FAB-MAP also performs very well (e.g. Oxford \cite{ds_city_center_new_college}), while algorithm Y is able to beat \mbox{FAB-MAP} by a large margin on another dataset that is particularly challenging for FAB-MAP (e.g. Nordland \cite{ds_nordland}).
This is not evidence that Y is better than X on any of the two datasets.

\section{Distinguishing properties of place recognition problems}
\label{sec:pr_props}

\begin{table*}[tb]
\centering
\caption{Distinguishing properties}
  \begin{tabular}{l p{4cm} p{5cm} p{5cm}}
  & \textbf{Property} & \textbf{(More) simple instances} & \textbf{(More) difficult instances}\\
  \hline
  \hline
  & \multicolumn{2}{l}{\textit{A - Knowledge level (Sec.~\ref{sec:pr_props_knowledge})}} \\
  \hline
  A1 & Are there additional sensors? & odometry and/or other sensors & (standard) vision only\\
  A2 & online vs. offline & database and query are known in advance & full online \\
  A3 & Are there unknown places during query? & for each query image there is a corresponding database image & exploration of potentially unseen places\\
  A4 & Is there additional knowledge available about any other aspect? & any additional knowledge or assumption can considerably simplify the problem & no knowledge about potential condition or viewpoint changes, trajectory properties, ... \\
  \hline
  \hline
  & \multicolumn{2}{l}{\textit{B - Viewpoint changes (Sec.~\ref{sec:pr_props_viewpoint})}} \\
  \hline
  B1 & amount of viewpoint change & small or no viewpoint changes & large viewpoint changes\\
  B2 & Is there a discrete set of places? & well distinguishable discrete places & a continuum of places\\
  B3 & degrees of freedom & limited degrees of freedom & full 6-DOF\\
  \hline
  \hline
  & \multicolumn{2}{l}{\textit{C - Intended output (Sec.~\ref{sec:pr_props_output})}} \\
  \hline
   C1 & single best matching or all & single best matching & requesting all matchings\\
   C2 & definite matchings or a list of candidates & a high number of potential candidates & definite matchings\\
   C3 & similarities or decisions & requesting similarities & requesting hard decisions\\
  \hline
  \hline
  & \multicolumn{2}{l}{\textit{D - Dataset scale (Sec.~\ref{sec:pr_props_scale})}} \\
  \hline
   D1 & number of images & small datasets & (very) large scale\\
   D2 & runtime requirements & ignoring runtime & hard real-time requirements \\
   D3 & storage limitations & ignoring memory consumption & requirement for small descriptors \\
  \hline
  \hline
  & \multicolumn{2}{l}{\textit{E - Domain (Sec.~\ref{sec:pr_props_domain})}} \\
  \hline
   E1 & type of environment & well defined and static environment with well distinguishable places& mixed (unknown) and/or dynamic environment with many similar looking places\\
   E2 &  application area/ vehicle & slow and restricted camera motion & fast and agile camera motion \\
  \hline
  \hline
  & \multicolumn{2}{l}{\textit{F - Appearance change (Sec.~\ref{sec:pr_props_appearance})}}\\
  \hline
   F1 & character/intensity of changes & (almost) constant appearance of places & severe appearance changes and bad illumination/visibility conditions\\
   F2 & Is there a discrete set of condition? & only two or a small number of discrete conditions & continuous conditions\\
   F3 & Can the condition change within a sequence? & no in-sequence changes & possibly in-sequence changes\\
   F4 & knowledge about conditions & complete knowledge (maybe also including training data) & no knowledge about potential and actual condition\\
  \hline
  \hline
  & \multicolumn{2}{l}{\textit{G - Spatio-temporal structure (Sec.~\ref{sec:pr_props_seq})}}\\
  \hline
   G1 & sequences or individual images & long sequences & individual images\\
   G2 & assumptions about velocities & constant (or even same) velocity in database and query & arbitrary velocities\\
   G3 & Are in-sequence loops possible? & no in-sequence loops & potentially multiple loops\\
   G4 & Are stops possible? & no stops & potentially arbitrarily long stops\\
  \hline
  \end{tabular}
\label{tab:dist_properties}
\end{table*}

The previous section demonstrated that we have to be very careful when comparing place recognition results across different experiments.
In this section, we will take a closer look at the underlying reasons.
We identified seven aspects of visual place recognition experiments whose specification can yield considerably different place recognition problem types with different levels of difficulty.
These aspects are listed in table~\ref{tab:dist_properties} together with examples for particularly difficult instances.

\subsection{Knowledge level}
\label{sec:pr_props_knowledge}

The amount of available knowledge in a place recognition algorithm can be very different.
This is property "A" in table~\ref{tab:dist_properties}. In the following, we will discuss four aspects: 

(A1) A straightforward aspect is additional sensor information, e.g. from other exteroceptive sensors beyond standard cameras, like stereo, RGB-D, IR or LiDAR, or proprioceptive odometry measures.
Additional sensors can considerably simplify the visual place recognition problem. For example, by allowing to measure previously unobservable quantities (e.g. depth with absolute scale) or providing advantages in low-illumination situations (e.g. LiDAR at night). Also related is additional information that can be derived from the (standard) images, but requires additional computational modules, e.g. visual odometry or semantic segmentation \cite{Neubert2021b}. The availability of these modules can create additional limitations, e.g. if no (good) semantic segmentation module for low-illumination conditions in natural environments is available.

(A2) The knowledge level can also vary along the spectrum of online vs. offline problem setups.
Along this spectrum, we basically distinguish how much of the database and query set is known in advance.
In full offline setups, database and query sets are completely known in advance. 
On the other end of the spectrum, we can face a full online setup: starting from an empty database, query images are presented once at a time and have to be matched to all previous query images (which then form the database for this query). This is the expected setup in a SLAM system.
A practically relevant setup in the middle of this spectrum is a database that is completely known in advance, but the query images are only presented during operating time (this can be expected in many practical localization setups).
Setups with previously known database and/or query images allow for extensive preprocessing, e.g. domain specific training~\cite{McManus2014} or standardization~\cite{Schubert2020}.

(A3) Another aspect of the knowledge level is whether there are potentially unknown places during the query run (i.e. places in the query set for which there are no images in the database set). This exploration or open-world property can considerably increase the difficulty to select criteria to establish a matching (i.e. we cannot simply choose the most similar database image).

(A4) In general, any additional knowledge about any of the other here presented aspects can have significant influence on the complexity of the problem. Therefore, it should be obligatory to explicitly list this required knowledge for newly presented approaches. For example, expected levels of viewpoint changes (Sec.~\ref{sec:pr_props_viewpoint}), knowledge about possible appearance conditions (Sec.~\ref{sec:pr_props_appearance}), or assumptions about camera trajectories (Sec.~\ref{sec:pr_props_seq}). 

\subsection{Amount of viewpoint changes}
\label{sec:pr_props_viewpoint}
In contrast to pose estimation, place recognition is an ill-defined problem, since answering the question whether two images show the same place or not requires additional information about the maximum allowed camera pose discrepancy. As said before, an image of the Eiffel tower and Notre-Dame cathedral might be considered showing different places (since there is no common image content) or showing the same place (Paris).

(B1) The amount of viewpoint change can have significant influence on the difficulty of the place recognition problem and the suitability of a particular approach.
For example, changing viewpoints challenge image descriptors that rely on constant positions of image features like the early ConvNet approaches to place recognition that used flattened feature map responses of AlexNet as descriptors \cite{Suenderhauf2015}.

(B2) In another direction, we can distinguish applications and datasets with a discrete set of  well separated places in the database (for example where each place is represented by a single image) and those with a continuous sequence of partially overlapping views of an area that is considered to contain several different places. The latter also poses particular challenges to the definition of ground truth (as was illustrated in Fig~\ref{fig:infl_gt_distance}).

(B3) Many mobile robotics place recognition datasets for changing environments contain only relatively small viewpoint changes since they are focused on ground vehicles, often in practical setups with roads or pathways that additionally limit the degrees of freedom of the camera movement.
Often there are only longitudinal, small lateral, and small rotational pose shifts.
Datasets with pixel-aligned images like Nordland~\cite{ds_nordland} can be used to systematically evaluate the influence of (synthetic) viewpoint changes.

\subsection{Intended output}
\label{sec:pr_props_output}
A very important distinguishing property is the intended output of a place recognition algorithm.
The basic pipeline from Sec.~\ref{sec:pr_basics} stated that the output is a set of associations between a query and database sets.
However, the characteristics of these associations for a query image can be significantly different. It can be:
\begin{enumerate}
 \item A single best matching.
 \item (C1) A list of \textit{all} database images that show the same place. This is typically required in online setups when it is not possible to preprocess the database to create a list of unique images per place or when we are interested in multiple matchings, e.g. to add more constraints for the map optimization during SLAM.
 \item (C2) A list of candidates that will be further refined in a verification step. Here, the intention is to have at least one true-positive matching among the candidates; the presence of additional false-positives is irrelevant (which distinguishes this case from the previous).
 \item (C3) Finally, many algorithms do not output a discrete set of associations but the similarity matrix $S$. The actual creation of place matchings is then left to further processing stages.
\end{enumerate}

The difficulty can significantly vary between these cases.
For example, the OPR \cite{Vysotska2016} approach builds on the assumption of single matches 
and can benefit from this assumption. However, it is also likely to fail if this assumption is violated in a dataset as shown in \cite{Schubert2021a}.

Reducing the output to the similarities instead of discrete matchings brushes the difficulty to make concrete decisions aside, e.g., to select a threshold on the descriptor similarity for a particular dataset. This is also hidden in the evaluation based on (the area under) precision-recall curves that are created based on a variety of thresholds. 
This also ignores the beneficial property of an algorithm to create large margins between true and false matchings or to allow the usage of the same threshold across many datasets (or procedures to determine the threshold for a dataset).

\subsection{Dataset scale and runtime/memory requirements}
\label{sec:pr_props_scale}

(D1) The number of expected database and query images can significantly vary.
The datasets that are typically used for evaluation in mobile robotics range from a few hundred images (e.g. GardensPoint Walking \cite{ds_gardenspoint_riverside}) to several thousand (e.g. Oxford Robot Car \cite{ds_robotcar}) and typically cover rather small areas like a campus or a few roads of a city. 

(D2, D3) Dependent on the practical application context of place recognition there can be limitations on the available time for each query (including real-time requirements) and limitations on the available storage space for the database.
As discussed in Sec.~\ref{sec:pr_basics} many place recognition approaches combine a descriptor computation stage with a matching stage.
For small scale datasets, the computational costs at the descriptor stage might dominate the overall computation costs.
However, the effort for computing descriptors depends linearly on the numbers $n$ and $m$ of images in the database and query sets ($O(n+m)$) but the costs for an exhaustive pairwise comparison grows significantly faster ($O(n \cdot m)$).
Therefore, (very) large-scale place recognition problems pose significantly different challenges not only for descriptiveness but also regarding runtime and memory consumption \cite{Neubert2021a, Schubert2021a}.

\subsection{Domain}
\label{sec:pr_props_domain}

The application domain has very large influence on many properties from table~\ref{tab:dist_properties}. Thus there is some overlap of this section with others. 

\begin{figure}[tb]
    \centering
    \includegraphics[width=1\linewidth]{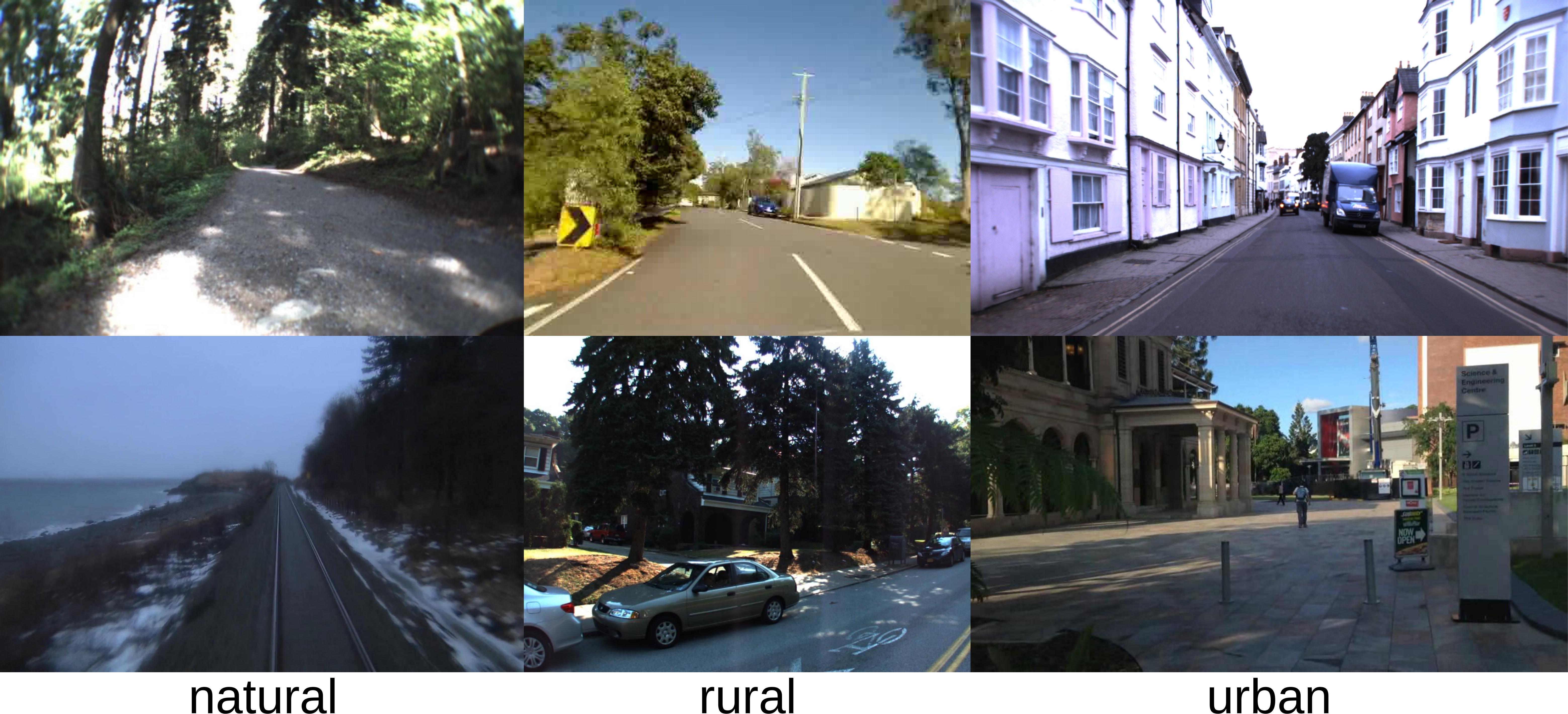}
    \caption{Example images for the most common types of environment in large-scale place recognition} 
    \label{fig:ex_type_env}
\end{figure}

(E1) One important aspect of the domain is the type of environment for place recognition.
Typical distinctions are indoor-outdoor or urban-rural-natural (see Fig.~\ref{fig:ex_type_env} for some example images).
Some environments tend to have different places that look very similar. This visual aliasing is particularly challenging for place recognition.
The environment also has important influence on the expected type and amounts of appearance changes, viewpoint change, and dynamic objects.
It is also particularly important when integrating additional semantic information (e.g. recognizing the type of a room) or when using learning-based methods like NetVLAD~\cite{netvlad} or DELF~\cite{delf} that were trained on datasets with images from one type of environment and could perform worse if applied in environments different to the training data.

(E2) Another important aspect of the domain is the application area and, strongly related, the device/vehicle that provides the (query) images. For example, place recognition in an automotive scenario exhibits less viewpoint change but more dynamic objects than an autonomous air vehicle that flies through a forest.
The application area is also expected to influence the criticality of false-positive matchings and the intended output as described before in Sec.~\ref{sec:pr_props_output}.

\subsection{Appearance changes}
\label{sec:pr_props_appearance}

Long-term operation of robots and long-term usage of vision-sensor-based maps create an increasing interest in place recognition despite severe appearance changes of the environments, e.g. due to lighting (e.g., dawn, sunny, dusk, night), weather (e.g., sun, rain, snow, storm), or seasonal changes (e.g., summer/winter, dry/wet season).
Of course, the intensity and type of appearance change has significant influence on the difficulty of a place recognition problem setup and on the selection of suitable approaches.
However, there are also other, less obvious aspects with potentially significant influence. 
Overall, we identified the following four aspects of appearance changes as distinguishing properties of place recognition problems:

(F1) \textit{Character and intensity of appearance change:}
The spectrum ranges from mild lighting changes to day-night or winter-summer changes.
Often, there is a trade-off between robustness towards appearance and viewpoint changes.
Methods like FAB-MAP~\cite{ds_city_center_new_college} have shown good performance for visual place recognition in environments with dynamic objects and slight changes in illumination.
In contrast, visual place recognition in \textit{changing} environments is still challenging and therefore subject of active research.
For example, the authors of \cite{Furgale2010} observed in their localization experiments that local SURF features~\cite{surf} fail under strong lighting changes.
Similarly, \cite{Neubert2015} concluded that the performance of the {SIFT-DoG-6} detector for place recognition drops with increasing severity of condition changes.

(F2) \textit{Is there a discrete set of conditions or a continuum?}
Most place recognition approaches for changing environments are evaluated on datasets where the database and query sequence have different conditions, but each has a single condition out of a discrete set of possible conditions.
This can be exploited, for example, in domain adaptation approaches that are able to translate images between conditions (e.g., \textit{ToDayGAN}~\cite{Anoosheh2019}) or by simple standardization approaches (see \cite{Schubert2020} for a discussion).
However, in many real long-term operation scenarios, we can expect a continuum of appearance conditions, for example, various types of ``rainy'', a smooth transition between day and night, and no clear distinction between a blue sky and overcast conditions.
Please refer to \cite{Schubert2020} for a more detailed discussion about continuous condition changes.

\begin{figure}[tb]
    \centering
    \includegraphics[width=0.39\linewidth]{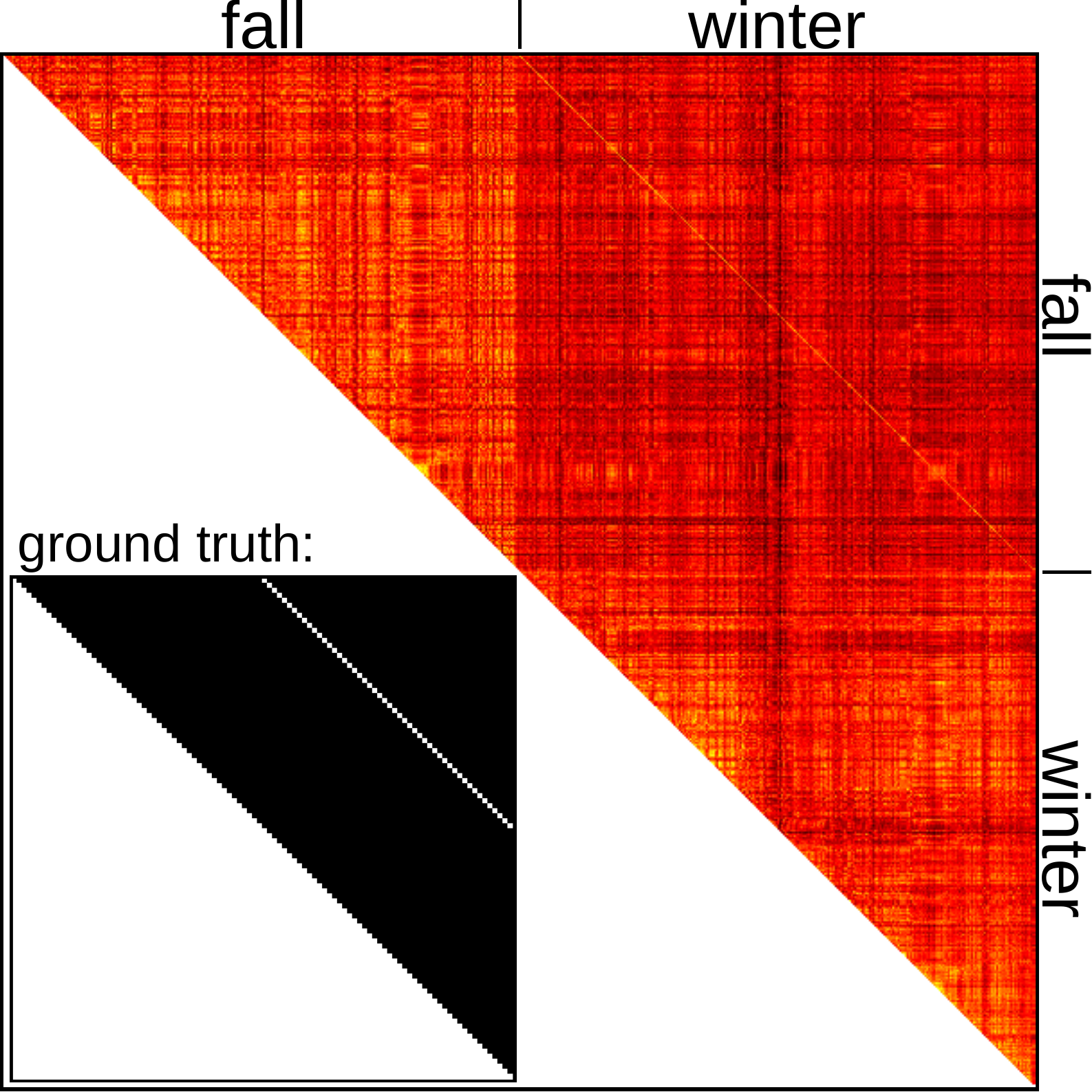}    
    \includegraphics[width=0.59\linewidth]{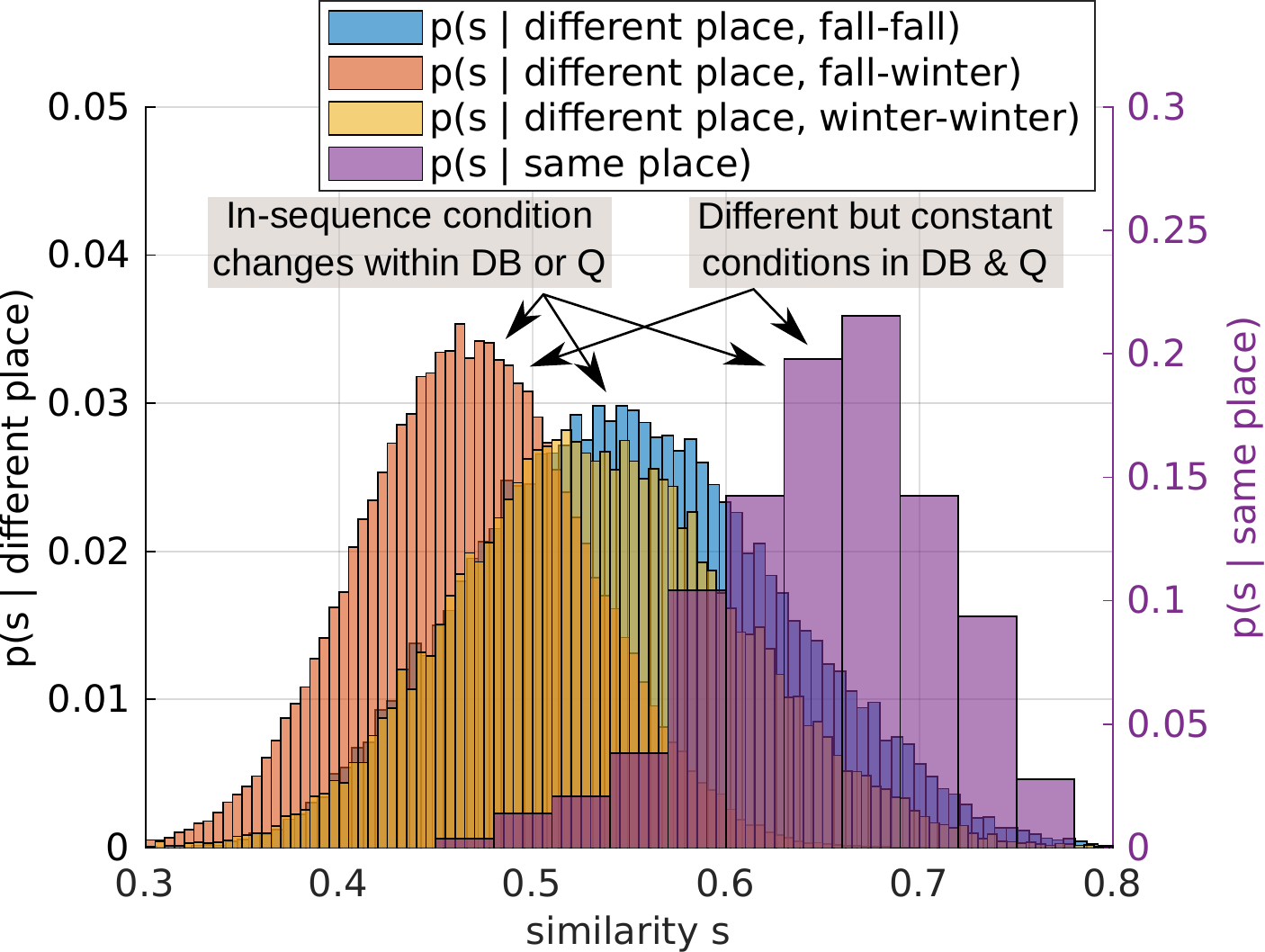}
    \caption{Descriptor similarities $s$ from pairwise AlexNet descriptor~\cite{Suenderhauf2015} comparisons $S$ and corresponding conditional probabilities $p(s|\cdot)$ from the Nordland dataset with in-sequence condition change from fall to winter. The ground-truth matchings (see bottom left) in fall-winter cannot be well distinguished from \textit{different} places in winter-winter and fall-fall.    
    }
    \label{fig:cont_cond}
\end{figure}

(F3) \textit{Can the condition (continuously) change within the database and/or query set?} 
Strongly related to the previous aspect is the question whether conditions can change \textit{within} the database or query sequence. 
As recently discussed in \cite{Schubert2020}, most approaches and datasets in the literature assume different but constant conditions within database and query, for example database is ``day'' and query is ``night''. But what if the query sequence was captured during the transition from day to night? 
Fig.~\ref{fig:cont_cond} uses the example of place recognition on Nordland using a combined ``fall-winter'' condition to illustrate the problem for the choice of the matching criteria: Different places under the same condition tend to look more similar than the same place under different conditions.

Fig.~\ref{fig:cont_cond} shows the corresponding pairwise image similarity matrix $S$ obtained from the AlexNet-conv3 descriptor~\cite{Suenderhauf2015} with ground truth, and conditional probabilities $p(s|\cdot)$ for all similarities $s$ in $S$.
Since we assume a single online run and compare each current image only to its predecessors, an upper triangular matrix is received.
After an initial traverse in fall without loops, the camera revisits all places once in winter.
Accordingly, high similarities (i.e. bright pixels) should only appear as single minor diagonal in \textit{fall-winter} image comparisons.
However, $S$ (Fig.~\ref{fig:cont_cond}, left) has high similarities (i.e. bright pixels) as well in \textit{fall-fall} and \textit{winter-winter} image comparisons, although these do not contain similarities of same places.

The histograms in Fig.~\ref{fig:cont_cond} (right) support this intuition: In place recognition experiments with different but constant conditions in database and query, we would merely observe the most left probabilities $p(s|\text{ different place, fall-winter})$ and most right probabilities $p(s|\text{ same place})$.
These two distributions have only a small overlap and are therefore well separable which leads to a high place recognition performance.
In contrast, in continuously changing environments we additionally obtain the probabilities $p(s|\text{ different place, fall-fall})$ and $p(s|\text{ different place, winter-winter})$ that have a much higher overlap with $p(s|\text{ same place})$ in Fig.~\ref{fig:cont_cond} (right).
As a consequence, same and different places are less separable which can, if not taken care of, lead to a significant performance drop.

(F4) \textit{Knowledge about possible conditions:}
Finally, for each of the above aspects, the level of knowledge is crucial. 
This includes knowledge about the principle problem setup (Which conditions might appear?), knowledge about the particular problem instance (What is the current condition?) as well as the availability of training data, e.g. to learn to detect the current condition, learn invariant representations, or to translate between different conditions.

\subsection{Spatio-temporal structure}
\label{sec:pr_props_seq}
\begin{figure}[tb]
    \centering
    \includegraphics[width=0.8\linewidth]{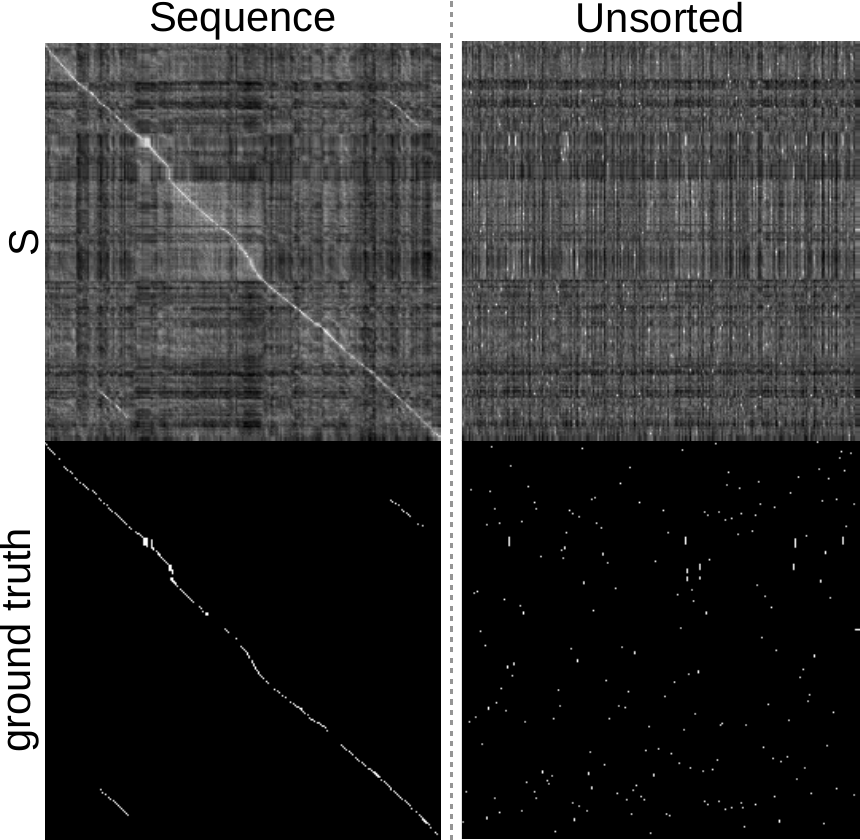}
    \caption{It is much easier to find all matchings in the similarity matrix $S$ of a temporally ordered sequence than in the unordered image set (part of StLucia 08:45).} 
    \label{fig:sequence}
\end{figure}

(G1) In contrast to a general image retrieval problem, for the visual place recognition problem considered here, images are typically recorded by a robot, car, or person equipped with a camera that is driving or walking along a trajectory in an environment.
Consequently, images in database and query might be acquired as spatio-temporal \textit{sequences}.
Consecutive images show either same or neighbouring places, and were likely recorded under the same or a very similar condition.
If we then compare images from a database and query, we likely receive a similarity matrix $S$ with one or multiple salient sequences of high similarities like in Fig.~\ref{fig:sequence} (left).
This is exploited by sequence-based place recognition approaches to improve the place recognition performance \cite{seqSLAM,Neubert2019,Naseer2014,Schubert2021b}.

\begin{figure}[tb]
    \centering
    \includegraphics[width=1\linewidth]{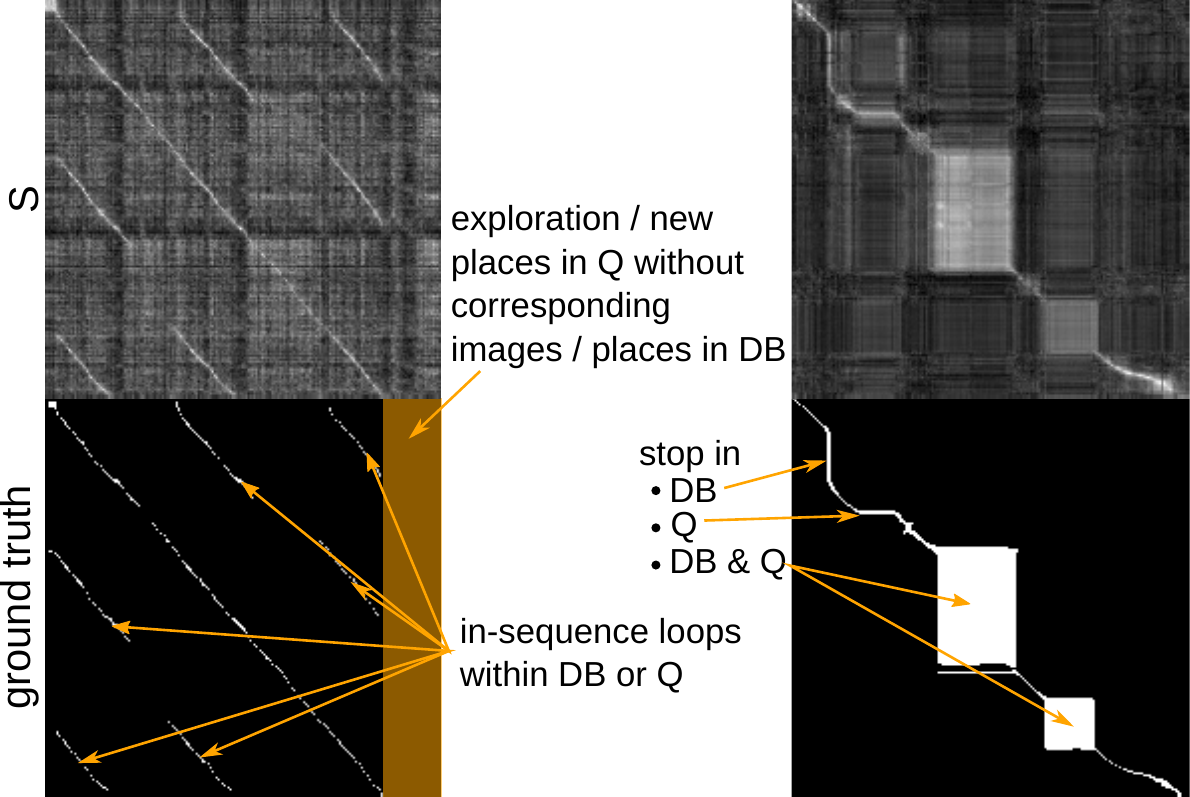}
    \caption{Typical patterns in pairwise similarity matrices $S\in\mathbb{R}^{|DB|\times|Q|}$ between the database (DB) and query (Q) set that are caused by the spatio-temporal structure of datasets.}
    \label{fig:loop_stops_overlap}
\end{figure}

However, dependent on the trajectories, velocities, and frame rates in the database and query sequences, the spatio-temporal structure (and thus the sequences in the similarity and ground-truth matrices) can significantly vary. 
Typical patterns are illustrated in Fig.~\ref{fig:loop_stops_overlap}:
\begin{enumerate}
    \item If the camera moves non-stop and never revisits a same place within database and query (but there are revisits between database and query), a single sequence similar to Fig.~\ref{fig:sequence} (left) can be observed.
    \item If places are revisited \textit{within} database or query (\textit{loops}), additional sequences with high similarities $s$ in $S$ appear (Fig.~\ref{fig:loop_stops_overlap}, left).
    \item If the camera temporarily \textit{stops} or moves very slowly on its trajectory, consecutive images show the same place in database or query.
    If the camera stops merely in the database and moves at the same place in the query, a vertical sequence of high similarities appears in $S$ (Fig.~\ref{fig:loop_stops_overlap}, right).
    Similarly, if the camera stops only in the query and moves at the same place in the database, a horizontal line of high similarities appears in $S$ (Fig.~\ref{fig:loop_stops_overlap}, right).
    And finally, if the camera stops at the same place in both database and query, a rectangle of high similarities in $S$ can be observed (Fig.~\ref{fig:loop_stops_overlap}, right).
    \item If new places are explored in the query, low similarities should be observed for all explored query images over all database images (Fig.~\ref{fig:loop_stops_overlap}, left).
\end{enumerate}

This is important since many sequence-based place recognition approaches make (implicit) assumptions about this structure and potentially fail if these assumptions are not met. Dependent on whether these assumption can by enforced by appropriate preprocessing or not (e.g. using visual odometry and/or image resampling), they might limit the general applicability of a place recognition approach. Typical assumptions about spatio-temporal structure are:
\begin{enumerate}
 \item (G2) Constant velocities in the database and query (or even the same velocity)
 \item (G3) No loops within the database sequence (and potentially also the query). 
 \item (G4) No stops in database and/or query.
 \item (cf. A3) No unseen places during query.
\end{enumerate}

\section{Discussion}
\label{sec:discussion}

The previous sections argued that place recognition experiments are barely comparable across papers and that there is a variety of properties that can change from one experiment to another. 
We think this can have severe implications for different involved parties: people who just want to use place recognition as a tool, researchers that look for open research questions, authors that want to create good papers on this topic, and also reviewers that have to identify problems in submitted papers.

\subsection{There is no one-fits-all solution (yet?)}
The large variety of properties of place recognition problems makes it very hard to design a single approach that meets all potential requirements. The ill-defined nature of the place recognition problem can create incompatible requirements. For example, one application might call for image matchings despite severe viewpoint changes, another might be much more restrictive on possible deviations of the camera pose.
From a more general perspective, as in many problems, the more specific we can tailor the solution to the particular problem instance, the better it can solve the problem.
Often, this tailoring is based on the introduction of additional assumptions on the previously discussed properties. For example, assumptions about possible camera motions  (Sec.~\ref{sec:pr_props_seq}) or potential appearance changes (Sec.~\ref{sec:pr_props_appearance}).
Introducing such bias in the algorithm can make it completely unsuited for problem instances where the assumptions are violated.

A promising approach to address this issue is to automatically determine the properties of the place recognition problem at hand and to select the appropriate approach. However, this seems to be largely unexplored terrain for place recognition.

\subsection{There are open research questions}
Although place recognition is an extensively studied problem, there are challenging open research questions. 
The sheer number of properties that can vary in the setup of a place recognition experiment makes the creation of a standard benchmark an important and challenging research task on its own.

The third column of table~\ref{tab:dist_properties} provides examples of (more) difficult instances of each property. In particular, problem setups, that combine multiple of these difficult properties are also likely to be open research topics. For example, a A1+B1+F1+G1-problem that is (standard) camera only (A1), with large  viewpoint changes (B1), potentially severe appearance changes (F1), and no spatio-temporal sequences (G1).

Examples for individual properties that challenge existing approaches are in-sequence appearance changes (F3) and finding \textit{all} database matchings for a query instead of a single match (C1). 
Of course, very bad illumination or visibility conditions (F1, heavy rain, snow, night-time, overexposure, etc.) are a challenge on their own and improvements on local and holistic image descriptors can still make a huge difference.

With a more general perspective, computational demands (D1+D2+D3) and power consumption can be critical on mobile devices like robots. For example, their limited resources can hamper the usage of powerful yet expensive approaches like pairwise local feature matching and usage of epipolar geometry for more than a few candidates per query. Resource-efficient usage of the representative power of local features in combination with geometric verification can also be a promising direction for future research.

As said before, based on the observation that additional assumptions can considerably simplify the problem and their violation can bungle a particular approach, research towards automatic testing of the validity of such assumptions can potentially have huge impact for practical applicability.

\subsection{Papers have to provide a lot of information about the problem and experimental setups}
As long as there is no such automatic adaptation, for practical application it is crucial to know the properties of the place recognition problem at hand, as well as the properties, assumptions, and limitations of the existing algorithmic approaches.
Basically, each paper that introduces a novel place recognition approach and/or provides experimental results should provide information about each of the properties from table~\ref{tab:dist_properties}.
This information is also essential to ensure repeatability of experiments and to increase the comparability between papers.

\subsection{Cherry picking is easy}

The previous Sec.~\ref{sec:pr_props} showed the variety of different place recognition problem types.
It is clear that different place recognition approaches might be differently well suited for different problem types.
For example, one image description approach might be more robust to appearance changes (e.g., AlexNet) and another more robust towards viewpoint changes (e.g., SIFT keypoints).
Table~\ref{tab:influence_dataset} demonstrated that the relative performance of NetVLAD and AlexNet can switch dependent on the evaluated dataset.
Given the large variety of existing datasets (cf. table~\ref{tab:datasets}), there is a good chance that for some new algorithms there is one or a few datasets where it can beat some other algorithms.

An important observation is that such different strengths of approaches are often also reflected in the design of experiments and evaluations of papers that propose new approaches.
However, this can potentially hide weaknesses of the approach or requirements for its applicability.

For sequence-based approaches, an often made (and sometimes hidden) simplification are restrictions on the trajectory of the camera, like assuming constant velocity between database and query sequences. Even if the origin velocities of the cameras might vary, some papers apply a resampling of the camera video stream to discrete images based on the ground-truth poses. This resampling can simplify the problem to merely searching for main or secondary diagonals in the pairwise image similarity matrix. Of course, the ground-truth pose knowledge is not available in practical applications which makes this approach flawed. 

Basically, all (more) simple instances of properties from table~\ref{tab:dist_properties} can be used to simplify the problem. For example, assuming that there are no unseen places during query (A3), aligned viewpoints (B1), only a single matching database image for each query (C1), knowledge about possible conditions (F4), and so on. 

To make this clear, there is nothing wrong in making those assumptions, as long as (1) they are clearly stated, (2) they are realistic in some (practical) scenario, and (3) the assumption is appropriately reflected in the experimental comparison to other approaches, in particular by comparing against methods that also exploit this simplification (if there are any).

An example for a violation of the last point was discussed in Sec.~\ref{sec:barely_comparable_prepost}: proposing a new descriptor for place recognition that includes a descriptor standardization step and then comparing against existing approaches without this standardization.

To reduce or prevent confusion about assumptions and properties, all additional assumptions, required knowledge, or pre/postprocessing steps should be made clear and justified for newly presented place recognition approaches. Moreover, evaluation should always be performed on a broad range of datasets, e.g. 10-20 sequence comparisons from multiple datasets. An indicator for suspect cherry picking can be when the selection of evaluated datasets changes from one experiment to the other within one paper without explanation.

\section{Conclusions}
\label{sec:conc}

We discussed that place recognition results are barely comparable across papers and that there is a variety of properties that can lead to considerably different types of place recognition problems.
We hope this increases the awareness of these aspects and supports according treatment in future research. 
The list of properties is not necessarily complete (nor is the chosen partitioning obligatory) and is likely to grow with future developments and further specialization in this field. 
If appropriate, we will consider an update of this list and any comments and further contributions are very much welcome.


\section*{Appendix A: Details on the basic place recognition pipeline}

\subsection{Input}
Input to a place recognition pipeline is a set $DB$ of $n$ database images and a set $Q$ of $m$ query images.
Depending on the application, $DB$ and $Q$ can be separate (e.g. global localization) or identical (e.g. SLAM).
In an \textit{offline} setup, all images in $Q$ are given in advance; in an \textit{online} setup, $Q$ is a growing set of images; a setup for \textit{delayed} place recognition uses a few future query images for performance improvements.
For \textit{realtime} requirements, new query images have to be processed in a certain amount of time.
$Q$ can solely contain images of same places as in $DB$ or also images of additional, unseen places (\textit{exploration}).

\subsection{Descriptor Computation}
The first step in the basic place recognition pipeline is the computation of some kind of descriptor from each database and query image.
Existing descriptors can generally be divided into local and holistic descriptors, which can both be either hand-designed or learned.
Hand-designed local descriptors like SIFT~\cite{sift}, SURF~\cite{surf}, or ORB~\cite{orb}  detect and describe local features in an image, and demonstrated good performance for place recognition in SLAM systems like FAB-MAP~\cite{ds_city_center_new_college} or ORB-SLAM2~\cite{orbslam2}.
However, it was shown several times in the literature that existing local feature detectors fail under environmental condition changes \cite[p.187]{ds_seasons, Furgale2010, densevlad, Neubert2015}.
Therefore, deep-learned local descriptors like DELF~\cite{delf} or D2-Net~\cite{Dusmanu2019} were proposed that achieve good performance despite condition changes.
There are also combined systems like EdgeBoxes~\cite{edgeboxes} or SP-Grid~\cite{Neubert2016}.
The drawback of local descriptors is the high computational effort when comparing two images for place recognition.

An often faster alternative are holistic descriptors that describe an image as a single (high-dimensional) vector.
In earlier work on place recognition, preprocessed pixel-values directly served as holistic descriptor \cite{seqSLAM,Lowry2014} or a single hand-designed descriptor for local features like BRIEF~\cite{brief} was computed from a full image \cite{briefgist}.
Later, the work \cite{densevlad} proposed DenseVLAD which computes local RootSIFT descriptors~\cite{rootsift} in a grid all over an image and combines them via VLAD~\cite{vlad} in a single holistic image descriptor.
The authors of ~\cite{Suenderhauf2015} proposed the usage of intermediate layer activations from Convolutional Neural Networks (CNNs) trained for object classification like AlexNet~\cite{alexnet}.
Descriptors like NetVLAD~\cite{netvlad} or DELG~\cite{delg} are CNNs directly trained for place recognition tasks.

\subsection{Computing the similarity between two images}
Given holistic or local descriptors, there are different ways to compute the similarity (or distance) between two images.
The mean absolute error \cite{seqSLAM, ds_surfer} or mean squared error \cite{Lowry2014} can be used to compare the preprocessed pixel-values of two images.
Cosine similarity (e.g., \cite{Suenderhauf2015}) or euclidean distance (e.g., \cite{netvlad}) is mainly used to compare holistic descriptors.

For local image descriptors, the similarity computation between two images is often divided into two steps: 1) find and validate local descriptor matches, 2) compute the image similarity based on the set of local features.
In the first step, local feature correspondences between two images can be found for example with a left-right check (also termed crosscheck or mutual matchings) \cite[Eq.4]{Neubert2015a} or with a geometric verification \cite{delf,Neubert2015}.
Additional aspects like the ratio between the closest and the next closest descriptors \cite[p.104]{Lowe2004} have been used as well.

Given the descriptor correspondences, the similarity can be computed in different ways.
In \cite[Eq.4]{Neubert2015a} the average similarity of all matching pairs is computed; \cite{ds_mapilary} adds a term for the local feature's box shape.
In \cite{delf}, the number of inliers after geometric verification is directly used as similarity score.

\subsection{Which images to compare between database and query}
Another important aspect that affects runtime and performance is which database images are compared to a current query.
Many approaches simply compare all images from the database with all query images (e.g., \cite{seqSLAM, Naseer2014}).
An alternative approach is to use a nearest neighbor search \cite{Dusmanu2019} with methods like Bags of Words \cite{bow} or different approximate nearest neighbor methods \cite{ann} like HNSW \cite{hnsw}.
A third way to reduce the number of image descriptor comparisons is the exploitation of spatio-temporal sequences within database and query (see Sec.~\ref{sec:pr_props_seq}) by selecting matching candidates from the database for the current query from the previous matching pairs \cite{Vysotska2016, Neubert2015b, Schubert2021a}.
In \cite{Vysotska2015}, available noisy GPS was used for candidate selection.

\subsection{Extensions of the basic place recognition pipeline}
There is a variety of existing extensions to the basic pipeline for performance improvements that build upon different assumptions.
The following methods from the literature are roughly sorted from input to output within the basic pipeline (Fig.~\ref{fig:infl_prepost}).
1)~Additional sensor data: If available, additional sensor data like odometry \cite{smart, Schubert2019} can be used.
2)~Image preprocessing: Some methods in the literature exploit prior knowledge about the condition changes.
If only the lighting condition changes due to time of day, illumination-invariant image conversion \cite{Alvarez2011,Ranganathan2013,Maddern2014} or shadow removal \cite{Corke2013,Shakeri2016,Ying2016} can be used.
In case of different but known condition changes, the conversion of query images into the database condition with appearance change prediction \cite{Neubert2013}, linear regression \cite{Lowry2016} or Generative Adversarial Neural Networks (GANs) \cite{Anoosheh2018,Anoosheh2019} have been proposed.
3)~Descriptor preprocessing: Particularly in an offline setup with all database and query images, a separate feature standardization of all descriptors for each condition \cite{Schubert2020} can be used to clearly improve the performance.
Without knowledge about when the conditions change, unsupervised learning based on clustering can be used \cite{Schubert2020}.
4)~Sequence-based methods: Since place recognition is an embedding of image retrieval into mobile robotics, spatio-temporal sequences within database and query can often be exploited.
MCN~\cite{Neubert2019} and Delta Descriptors~\cite{Garg2020} encode multiple consecutive descriptors into a new descriptor for each image.
Sequence-based methods like SeqSLAM~\cite{seqSLAM}, ABLE~\cite{able}, HMM and methods based on Flow Networks~\cite{Naseer2014,Vysotska2017} postprocess the similarity matrix $S\in\mathbb{R}^{|DB|\times |Q|}$.
5)~General postprocessing of $S$: Unsupervised learning methods based on PCA \cite{dr, Lowry2016} have been used to improve the pairwise image similarities \cite{Schubert2020}.

\subsection{Place recognition evaluation metrics}
\label{sec:literature_metrics} 

Place recognition evaluation typically builds upon knowledge of ground-truth place associations. For most available datasets this information is either given as a list of image pairs from $DB$ and $Q$ that show the same place or by an associated pose (e.g. based on GNSS) for each image. Thresholds on the (lateral, longitudinal, and rotational) distance between poses can then be used to compute a list of ground-truth matchings between $DB$ and $Q$. 
The assumed binary nature of this ground-truth information is an essential property of the ill-defined place recognition problem. This and the importance of the actual choice of the above pose thresholds is discussed in Sec.~\ref{sec:barely_comparable_gt}. 
The binary matrix $GT \in \{0,1\}^{n \times m}$ of the same size as $S$ indicates whether a pair of database and query image show the same place or not.

An often used basic evaluation pipeline is to take the similarity matrix $S$ as the output of the place recognition approach and to apply a series of varying thresholds $t_k$ with
\begin{align}
    t_k = [\min(S), \min(S)+\epsilon, \min(S)+2\epsilon, \ldots, \max(S)]
\end{align}
on $S$ in order to obtain binary matching decisions. 

Although the actual choice of the threshold is practically very important, this allows to evaluate the quality of the similarities independent of the quality of a particular threshold.
A low threshold $t_k$ is likely to create many image pairs which are actually correct (true positives $TP_k$) but also potentially classifies many image pairs as matchings which actually do not show the same place (false positives $FP_k$).
A high threshold in turn helps to prevent false positives but increases the risk of also missing database images that show the same place as the query (false negatives $FN_k$).
Given the binary matching decisions, the ground truth can be used to count these numbers \cite{Davis2006}:
\begin{align}
TP_k &= | (S\ge t_k) \land GT | \\
FP_k &= | (S\ge t_k) \land \lnot GT | \\
FN_k &= | (S < t_k) \land GT |
\end{align}
These values can then be used to compute the k-th point on a precision-recall-curve (PR-curve) by
\begin{align}
 \text{Precision: } P_k &= TP_k / (TP_k+FP_k) \\
 \text{Recall: } R_k &= TP_k / (TP_k+FN_k)
\end{align}
Precision-recall-curves are finally defined as the precision as a function of the recall with
\begin{align}
    P_k = f(R_k)
\end{align}
and serve as a widely used performance metric in the literature on place recognition.
Note that ROC-curves are a related performance measure but are unsuited due to the imbalance between matching and non-matching places \cite{Davis2006}.
While precision-recall curves provide detailed information about the performance of an algorithm, they require much space for visualization.
Therefore, they are not suited very well for an evaluation over many datasets.

To compress the information into a single number, different approaches have been used:
\begin{itemize}
\item The area under the precision-recall curve (AUC)
\begin{align}
    AUC = \int_0^1 P(R) dR
\end{align}

\item Maximum F1 score, which is the maximum harmonic mean of $P_k$ and $R_k$ \cite[p.781]{EncML} with
\begin{align}
    \text{maximum F1 score} = \max_k\frac{2\cdot P_k \cdot R_k}{P_k + R_k}
\end{align}
While AUC contains information about the full precision-recall curve, the maximum F1 score merely represents a single point on the precision-recall curve that represents a potentially good compromise between high precision and high recall.

\item Maximum recall at $100\%$ precision (Recall@100\%Precision).
This has later been modified to Extended Precision \cite{Ferrarini2020} to address the problem of Recall@100\%Precision, if the precision never hits $P=100\%$.
The Recall@100\%Precision was invented, because $100\%$ precision ensures solely correct loop closure in SLAM.
However, since the advent of robust pose graph optimization techniques around 2012 \cite{Suenderhauf2012}, avoiding wrong loop closures became less important.

\end{itemize}

An somewhat complementary metric is Recall@K which counts for which proportion of query images with true-matchings, one of the true-matchings is actually in the K highest rated database images.
It expresses the performance of a place recognition pipeline for candidate selection for global localization.

In addition to evaluation of place recognition performance, runtime and memory consumption are likely practically very important aspects.

\section*{Appendix B: Existing place recognition benchmarks}

\subsubsection{The Visual Localization Benchmark \cite{Sattler2018}}\label{sec:vis_loc_bm}

This benchmark\footnote{The Visual Localization Benchmark: \url{https://www.visuallocalization.net/}} was designed to evaluate and compare approaches for visual localization in changing environments.
The task is to compute for each query image a transformation estimation ($x$, $y$, $\theta$) that is then compared to three maximum-pose-error thresholds \textit{hard} (e.g., (0.25m, 2$^\circ$)), \textit{middle} (e.g., (0.5m, 5$^\circ$)), and \textit{easy} (e.g., (5m, 10$^\circ$)).
Their benchmark metric is described as: \textit{``the percentage of query images which where localized within three given translation and rotation thresholds''}.
The used datasets with the roughest threshold could potentially be used for place recognition.

Approaches that show very good visual localization performance use place recognition techniques for candidate selection before conducting expensive transformation estimation.
For example the winner of the \textit{CVPR'20 Localization Challenge}\footnote{CVPR'20 Localization Challenge: \url{https://sites.google.com/view/vislocslamcvpr2020/localization-challenges}} used NetVLAD~\cite{netvlad} for visual place recognition and candidate selection before feature computation and matching\footnote{Winner of the CVPR'20 Localization Challenge: \url{https://github.com/cvg/Hierarchical-Localization}}.
Other methods that performed very well on this benchmark like D2-Net~\cite{Dusmanu2019} or ONavi~\cite{Fan2020} used NetVLAD~\cite{netvlad} or DELF~\cite{delf}, respectively, for place recognition.

While the Visual Localization Benchmark is well suited for visual localization in changing environments, it does not consider aspects due to the chosen datasets or metric that are important for visual place recognition.
For example, the localization benchmark
\begin{enumerate}
    \item does not investigate the influence of condition change without viewpoint changes and vice versa.
    \item does not consider how condition changes \textit{within} database or query affect methods.
    \item evaluates only the global position of each query image. In contrast, in a SLAM setup, the goal of place recognition might be to find all matchings between query and database images. 
    On the one hand, this allows that query images need not to have a counterpart in the database and evaluates the ability to discover new unseen places. On the other hand, some place recognition algorithms exploit the knowledge that a query image has only a single matching database image. By evaluating all similarities between a query image and all database images, such assumptions can be revealed. These aspects are important for partial overlap between database and query as well as for loops and stops within database or query.
\end{enumerate}

\subsubsection{The VPRiCE Challenge 2015}

This challenge\footnote{The VPRiCE Challenge 2015: \url{https://roboticvision.atlassian.net/wiki/spaces/PUB/pages/14188617/The+VPRiCE+Challenge+2015+Visual+Place+Recognition+in+Changing+Environments}} was designed for a CVPR'15 workshop to benchmark and compare several submissions of place recognition algorithms.
The benchmark is a concatenation of sequences from the datasets Nordland, Mapillary and QUT campus and city into one database and query set.

The results of an algorithm had to be submitted as actual image matches (e.g., ``query 10 belongs to database 100''); i.e. the loop closure detection but not the place recognition performance is measured.
Further, the submitted file must contain only two columns with a constantly increasing query image number. That means, there must not be loops or stops in the database (that have an overlap with the query).

\subsubsection{All-Day Visual Place Recognition Benchmark \cite{ds_all_day}}
This benchmark\footnote{All-Day Visual Place Recognition Benchmark: \url{https://soonminhwang.github.io/publication/2015cvprwall-day-place-recognition/}} was proposed during the CVPR'15 workshop on visual place recognition in changing environments. The used datasets were recorded with a car equipped with RGB and thermal cameras as well as GPS and IMU around the Korea Advanced Institute of Science and Technology (KAIST) campus. While the benchmark may be interesting if thermal camera data is required, it covers merely a range of illumination changes in a single environment and is therefore not well suited for a comprehensive evaluation of visual place recognition algorithms over a variety of conditions and environments.

\subsubsection{Segway DRIVE Benchmark \cite{Huai2019}}
This benchmark\footnote{Segway DRIVE Benchmark: \url{http://drive.segwayrobotics.com/}} consists of several datasets collected by a fleet of delivery robots, and is designed for benchmarking place recognition and SLAM algorithms in indoor environments. Therefore, it covers appearance changes due to dynamic objects like people, long-term structural changes (e.g., removed furnitures), and illumination changes. Accordingly, it is unsuited as a benchmark that covers many kinds of condition changes in different large-scale environments.

\subsubsection{A realistic benchmark for visual indoor place recognition~\cite{Pronobis2010}}
Similarly to the Segway DRIVE Benchmark, the datasets of this benchmark were recorded in an indoor environment with two mobile robots and a tripod-camera, and covers dynamic objects, long-term changes and different lighting conditions. Due to the single environments and the restricted number of conditions, this dataset is unsuited as a large-scale, multi-condition benchmark.

\end{document}